\documentclass[11pt]{article}

\usepackage[preprint]{acl}

\usepackage{times}
\usepackage{latexsym}
\usepackage{times}
\usepackage{latexsym}
\usepackage{amsmath}
\usepackage{amssymb}
\usepackage{tcolorbox}
\usepackage{xcolor}
\definecolor{chineseblack}{HTML}{141414}
\definecolor{lightbluecobalt}{HTML}{88ACE0}
\usepackage{tikz}
\usetikzlibrary{arrows.meta, positioning}

\usepackage[T1]{fontenc}

\usepackage[utf8]{inputenc}

\usepackage{microtype}

\usepackage{inconsolata}

\usepackage{graphicx}

\title{Different Facets of Verbalised Overconfidence: \\ an Interpretability Study} 

\author{Davide Mazzaccara \\
CIMeC, University of Trento\\
\texttt{davide.mazzaccara@unitn.it}\\\And
Leonardo Bertolazzi\\
DISI, University of Trento\\
\texttt{leonardo.bertolazzi@unitn.it}\\\AND
Raffaella Bernardi\\
Free University of Bozen-Bolzano\\
\texttt{raffaella.bernardi@unibz.it}}

\begin{document}
\maketitle
\begin{abstract}
Large language models tend to overconfidence, giving assertive answers when the evidence suggests hedging or abstention. Using controlled reasoning scenarios that manipulate logical necessity and possibility, we study this behavior in Qwen3-4B, across three ways to express uncertainty: verbal epistemic markers, abstention, and numeric confidence scores. Our results confirm this tendency toward overconfidence, particularly when the model is prompted to output a numeric confidence score.
At the interpretability level, we propose a method that differentially identifies transcoder features responsible for uncertainty and certainty. Our analysis reveals Qwen3-4B’s default mechanism favors certainty generation through a broad coalition of shared features, while uncertainty is implemented as a sparse override mediated by a small set of dedicated features. Intervening on these uncertainty features both causally proves this imbalance underlying overconfidence and also mitigate overconfident errors. The same set of features generalise across the three uncertainty-expression settings, languages, and an out-of-distribution modality task.
\end{abstract}

\section{Introduction}
Large Language Models are increasingly permeating a wide range of practical domains across society, including critical scenarios like medical or intelligence operations. Despite their strong performance in generating appropriate outputs, their ability to accurately express uncertainty about those outputs remains a significant challenge \citep{xiong2024can, testoni-calixto-2026-mind}. A consolidated finding in the community is that Language Models tend to be overconfident \citep{zhou-etal-2024-relying}, producing assertive statements in situations where the evidence would suggest hedging or abstention. Previous studies, however, treat abstention as a different issue compared to verbalised confidence \cite{wen-etal-2025-know}, conflating the verbalised confidence expressed through verbal epistemic markers and numeric confidence scores \cite{zhou-etal-2023-navigating}. In this study, instead, we try a comprehensive approach comparing these three expressions of verbalised uncertainty: verbal epistemic markers, abstention and numeric confidence scores.


Controlled stories are an useful tool to study uncertainty expressions in language models. Each story reports some evidence and a question related to the story. If the evidence is enough to logically derive a yes/no answer, the model should convey certainty, otherwise uncertainty. The two samples below illustrate the verbal setting: certainty is expressed through the assertive \emph{is}, uncertainty through the epistemic marker \emph{might}. Uncertainty could be expressed also through abstention (\lq I don't know\rq), or with a low score on a scale from 1 (uncertain) to 5 (certain) about the answer.

\begin{tcolorbox}[
    colback=white,
    colframe=chineseblack,
    colbacktitle=chineseblack,
    coltitle=white,
    fonttitle=\small\bfseries,
    title=Certainty Scenario,
    halign title=center,
    boxrule=1.2pt,
    arc=5mm,
    left=2mm,
    right=2mm,
    top=1mm,
    bottom=1mm,    
    after upper={\vspace{-1mm}},
    before lower={\vspace{-2mm}}
]

\footnotesize
A pencil is in the red, the white or the blue box.
It is known the pencil is not in the red nor in the white box.

\medskip
Is the pencil in the blue box?

\tcblower

\footnotesize
It \emph{is} in the blue box.

\end{tcolorbox}

\begin{tcolorbox}[
    colback=white,
    colframe=lightbluecobalt,
    colbacktitle=white,
    coltitle=lightbluecobalt,
    fonttitle=\small\bfseries,
    title=Uncertainty Scenario,
    halign title=center,
    boxrule=1.8pt,
    arc=5mm,
    left=2mm,
    right=2mm,
    top=1mm,
    bottom=1mm,
    after upper={\vspace{-1mm}},
    before lower={\vspace{-2mm}}
]

\footnotesize
A pencil is in the red, the white or the blue box.
It is known the pencil is not in the white box.

\medskip
Is the pencil in the blue box?

\tcblower

\footnotesize
It \emph{might} be in the blue box.

\end{tcolorbox}
\begin{figure*}[h]
  \centering
  \includegraphics[width=\linewidth]{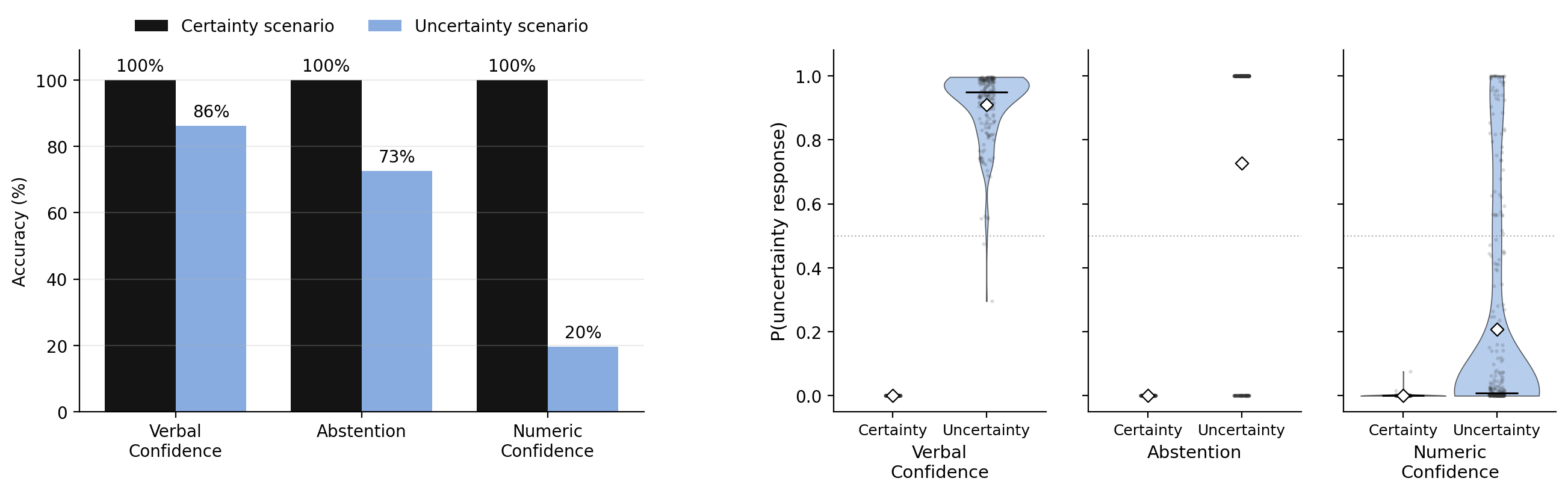}
  \caption{Qwen3-4B overconfidence in the three settings. Left: accuracy in the certainty and uncertainty scenarios. Right: distribution of the probability on the uncertainty-side response per scenario: in the verbal confidence setting is the summed mass for the observed epistemic tokens $P(\textit{might},\textit{could})$, in the abstention is the abstention rate, in the numeric confidence is $P(\text{score}<5)$. The diamond marks the mean.}
  \label{fig:overconf_channels}
  \label{fig:confidence_violins}
\end{figure*}
The primary purpose of this study is to better understand verbal overconfidence. Using Qwen3-4B model \citep{yang2025qwen3technicalreport} on controlled stories, we try to answer the following research questions:
\begin{itemize}
    \item \textbf{RQ1}: Does the model show overconfidence behavior in verbal confidence, abstention and numeric confidence scores?
    \item \textbf{RQ2}: What is the model's mechanism to express certainty vs. uncertainty?
\end{itemize}


\section{Experiment 1}\label{sec:experiment}
We test Qwen3-4B uncertainty behavior in the settings of verbal epistemic markers, abstention and numeric confidence score, using controlled stories. The research question is to observe if the overconfidence behavior emerges, along with possible differences among the three settings. The controlled stories are the ones by \citet{li-etal-2025-representations}, reframing its multiple-choice setting as an open-ended question-answering task. The dataset counts 450 pairs of premises and questions, 300 uncertainty scenarios and 150 certainty ones, spanning different templates. The dataset is tested in all three uncertainty-expressions formats (the prompts are in  Appendix~\ref{ap:dataset_prompts}). Accuracy is computed as the fraction of responses whose expressed (un)certainty matches the scenario. An overconfidence behavior in this context would be certainty answers in uncertainty scenarios. To evaluate accuracy, we need to evalute models' outputs as certain or uncertain, depending on the setting. For the verbal markers setting, given the wide range of expressions the model could produce, an external model as a judge (\href{https://www.anthropic.com/claude/haiku}{Claude Haiku}) has been employed to label the outputs as certain or uncertain. In the abstention setting, an uncertain label contains the abstention (\lq I don't know\rq), otherwise it is considered certain. In the numeric confidence score, the certainty label is '5', below it the model output is considered uncertain. 

\subsection{Overconfidence}\label{sec:elicitation}
As shown in Figure~\ref{fig:overconf_channels} (left), in the certainty scenarios the model consistently express certainty, whereas in uncertainty scenarios it incorrectly expresses certainty in $14\%$, $27\%$, and $80\%$ of cases in the verbal, abstention, and numeric settings, respectively. This tendency towards overconfidence is particularly pronounced in the numeric confidence setting, i.e., the model tends to state '5' (certain) in all scenarios. Figure~\ref{fig:confidence_violins} (right) further supports this observation, showing that most of the (mean) probability mass is on the '5' (certain) token even in uncertainty scenarios. The probability mass is near zero in certainty scenarios across all settings, and in uncertainty scenarios it concentrates near one for the verbal setting, splits between the two extremes for abstention. This imbalance between the distributions confirms overconfidence at the token probability level. Overall, the model exhibits clear overconfidence behaviours regardless of how uncertainty is elicited. An additional observation is that this tendency is particularly strong when uncertainty is expressed numerically, suggesting that numeric confidence scores are especially unreliable.


\section{Experiment 2}\label{sec:experiment2}

\begin{figure*}[t]
  \centering
  \includegraphics[width=\linewidth]{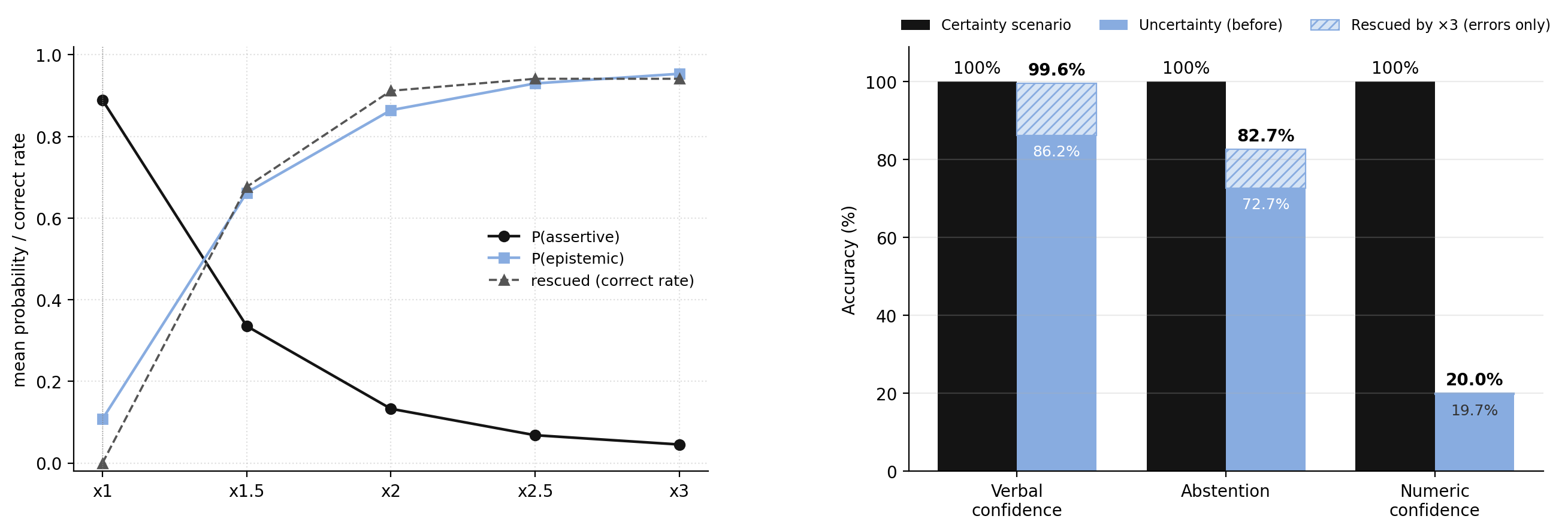}
  \caption{Error intervention on the top-20 uncertainty features for errors (uncertainty scenarios with certainty outputs). Left: graded effect of the boost multiplier on the error samples; $P(\textit{assertive})$ decays while $P(\textit{epistemic})$ and the rescued correct rate rise, reaching $\approx\!94\%$ from $\times 2.5$ on. Right: per-setting accuracy under the $\times 3$ boost.}
  \label{fig:intervention}
  \label{fig:rescue_channels}
\end{figure*}
\subsection{Methodology}
\label{sec:methodology}

Following the established practice in mechanistic interpretability of localizing units by contrastive selectivity and then validating them causally \citep{alkhamissi-etal-2025-llm}, we identify the transcoder features distinguishing certainty from uncertainty in Qwen3-4B in two complementary stages, run with \texttt{circuit-tracer} \cite{hanna-etal-2025-circuit} on the concept-level middle layers (20--29) over the correct samples: \emph{selectivity}---which features fire more for one condition than the other---and \emph{causal contribution}---which features actually push the next token toward the certain or uncertain expression. The two answer different questions: a feature can be selective without moving the output, or move the output without being condition-selective, so we keep both. For selectivity we contrast each feature's per-sample mean activation between conditions with Cohen's~$d$ (with Mann--Whitney~$U$ and Benjamini--Hochberg FDR at $\alpha=0.05$, suited to sparse, non-normal activations across many simultaneous tests), tagging features \emph{certainty-specific} ($d>0.2$), \emph{uncertainty-specific} ($d<-0.2$), or \emph{shared}. For causal contribution, on the top 500 selective features we run one attribution per sample toward $\mathrm{mean}(\mathrm{unembed}(\text{certainty}))-\mathrm{mean}(\mathrm{unembed}(\text{uncertainty}))$ (certain: \emph{is}, \emph{did}; uncertain: \emph{might}, \emph{could}), scaled by the realized $|P(\text{certain})-P(\text{uncertain})|$ so the score reflects the actual push on the logit, and record each feature's signed multi-hop influence on that target as its $\mathrm{dla\_score}$. Averaging the ranks on $|d|$ and $|\mathrm{dla\_score}|$ keeps features that are both selective and causally effective. Features selective for one condition but pushing the logit the other way are demoted and excluded from intervention.

\subsection{Default Mechanism vs.\ sparse override}
\label{sec:circuits}

The split between selectivity and causal contribution is not just methodological: it mirrors how the model actually produces certainty and uncertainty, and it is the source of overconfidence. At the \emph{activation} level the two conditions look alike---shared features dominate the mass in both (Figure~\ref{fig:error_profile}), so selectivity alone cannot tell them apart---but at the \emph{logit} level they diverge sharply: certainty-specific features contribute almost nothing to the certainty token, whereas uncertainty-specific features supply the bulk of the push to the uncertainty token. Certainty therefore has no dedicated subnetwork reaching the logit; it is the network's \emph{default}, carried by a broad coalition of shared middle-layer features. Uncertainty is its \emph{sparse override}: a small, dedicated group of uncertainty-specific features that takes over the push only when the prompt licenses it---broad and shallow for certainty, steep and concentrated for uncertainty within the same 20--29 band. Only the causal view singles out the features that actually drive uncertainty, which is why we rely on it to select intervention targets; the same asymmetry also explains overconfidence---certainty is the fall-through, so whenever the uncertainty override under-fires the model defaults to a confident answer, as the interventions of \S\ref{sec:intervention} confirm.


\subsection{Verbal, Abstention and Numeric Uncertainty}
\label{sec:intervention}

To establish that the features identified above \emph{cause} the model's certainty or uncertainty rather than merely correlating with it, we manipulate their activations during generation, boosting or suppressing the top-ranked certainty and uncertainty features, measuring the effect on the generated text. Boosting the top-20 uncertainty features 
at generation time on the 34 uncertainty error samples produces a graded rescue: none at $1\times$, $23/34$ at $1.5\times$, and $\approx\!94\%$ ($32/34$) from $2.5\times$ on, with $P(\text{uncertain})$ rising from $0.11$ to $0.95$ (Fig.~\ref{fig:intervention}). The converse ablation confirms the asymmetry: suppressing the same top-20 uncertainty-specific features on \emph{correct} uncertainty samples drives uncertainty expression from $99\%$ correct down to $0\%$, whereas the symmetric suppression of the top-20 certainty-specific features on certainty samples leaves certainty at $100\%$, and random-feature baselines are flat in both conditions (Appendix~\ref{ap:ablation}).

The same top-20 uncertainty features, boosted $\times 3$ on error cases only, generalize to the abstention setting: $30$ of the $82$ committed-answer errors flip to ``I don't know'', lifting abstention accuracy from $72.7\%$ to $82.7\%$ (Fig.~\ref{fig:rescue_channels}). The features that drive \emph{might}/\emph{could} thus also drive abstention, they encode uncertainty abstractly enough to surface in a different lexical form. On the numeric setting the same boost has a monotonic but negligible effect: the expected score drifts down monotonically with the multiplier, yet at $\times 3$ only $1$ of the $241$ over-confident errors crosses below $5$ ($19.7\% \to 20.0\%$, Fig.~\ref{fig:rescue_channels}). The score-$5$ probability mass is already saturated (mean $P(5)=0.95$ on the error cases), so the features cannot move the prediction. These uncertainty features seems to generalise in a multilingual dimension, as studied in Appendix \ref{ap:multilingual}.

\subsection{Uncertainty Features' Generalization}

\begin{figure}[t]
  \centering
  \includegraphics[width=\linewidth]{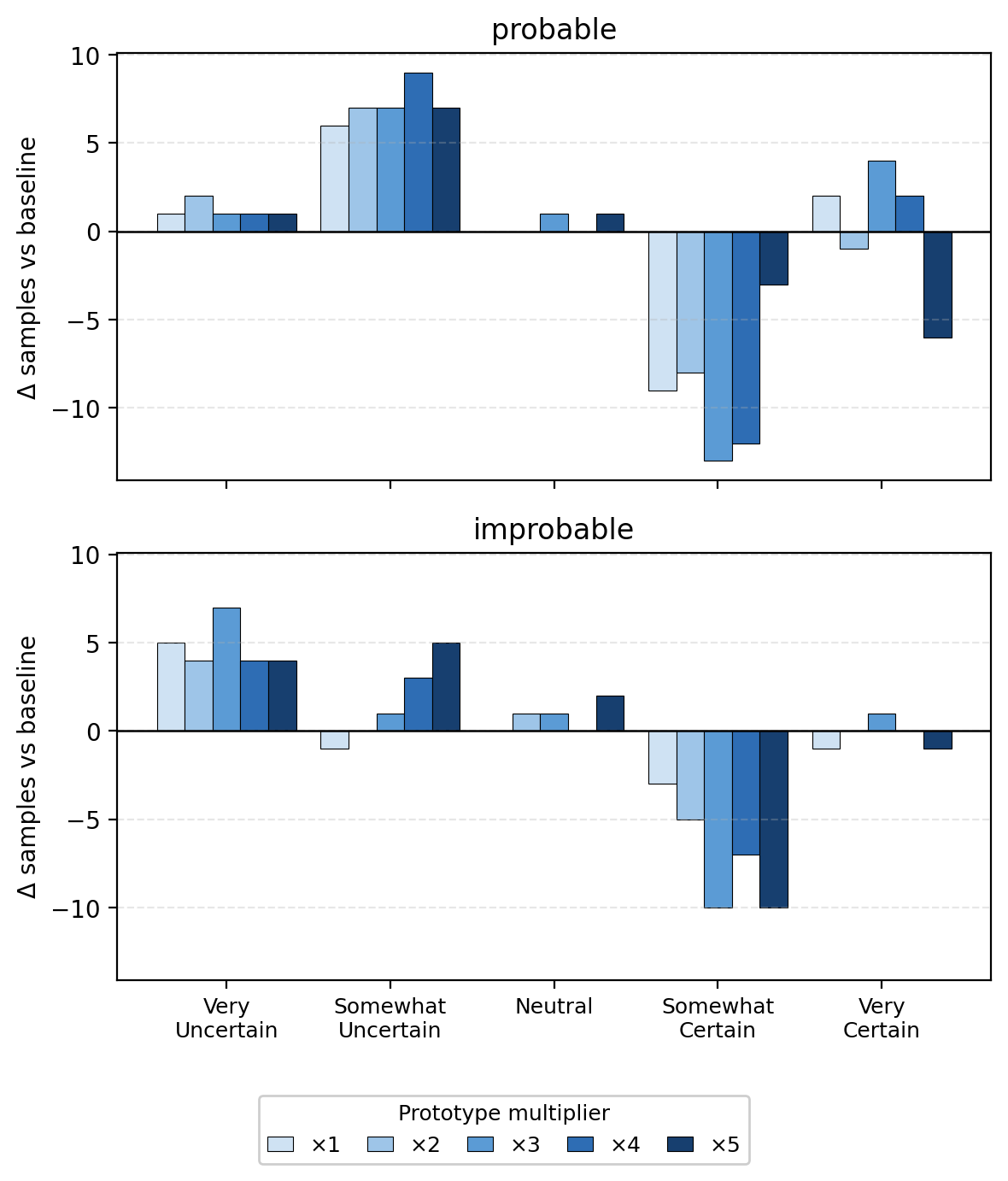}
  \caption{Generalization to the modality task. Change in the number of responses per certainty label under the prototype-value boost: from the baseline ($\times1$) to $\times5$. 
  }
  \label{fig:modality_delta}
\end{figure}

We test the uncertainty features for verbalised uncertainty by applying them to the modality task of \citet{lepori2026is}.
Our focus is on the modal conditions of \emph{probable}/\emph{improbable}, with 70 samples per condition. We clamp the same top-20 uncertainty features to the activations they have in the controlled stories' dataset, scaled $\times1$ to $\times5$ for generating with Qwen3-4B. The outputs are then rated by Claude Haiku, using a verbal scale from \emph{Very Uncertain} to \emph{Very Certain}. Figure~\ref{fig:modality_delta} reports, per certainty label, the change in the number of responses relative to the no-intervention baseline (bars above zero gained responses, below zero lost them; shading runs light to dark for $\times1$ to $\times5$). The boost moves mass out of \emph{Somewhat Certain} into the uncertain labels in both conditions, but asymmetrically: on \emph{probable} items the shift is confined to the \emph{Somewhat Certain}$\rightarrow$\emph{Somewhat Uncertain} edge, leaves \emph{Very Certain} intact, and saturates immediately (flat across $\times1$--$\times5$); on \emph{improbable} items it scales with the boost, draining \emph{Somewhat Certain} progressively into both \emph{Somewhat} and \emph{Very Uncertain}. The generalization of these features confirms the linear dimension of uncertainty expression recently found by \citet{ji-etal-2025-calibrating}. Uncertainty surfaces in several forms with intervention. Grammatically it weakens the modal verb, for \emph{``Could someone start a fire using kindling?''}, from ``someone \emph{can} start a fire'' to ``someone \emph{could} start a fire''. Lexically, for \emph{``Could someone chop a carrot using a sword?''}, ``it's \emph{certainly possible}'' becomes ``an \emph{unusual and impractical} scenario''. In some case intervention involves a rephraming of the answer, like for \emph{``Could someone eat soup using a bucket?''}, ``Yes, someone could technically\ldots'' becomes ``the question \ldots is a bit of a \emph{playful or hypothetical} one''.

\section{Conclusions}
In this study, we propose a way of looking at verbalised uncertainty, comparing and connecting verbal epistemic markers, abstention, and numeric confidence scores. The behavioral experiment observes the tendency towards overconfidence for the model in all these settings, finding the numeric confidence expression being the most prone to overconfidence. Secondly, we provide a method to identify transcoders' features for uncertainty and certainty verbal expressions, observing a particular mechanism in Qwen3-4B. The model seems to have a default assertive mechanism that is selectively overridden by uncertainty features, leading to verbal epistemic markers. Finally, via causal intervention we confirm the relation between the verbal uncertainty features an the other ways of expressing uncertainty. Their effects generalise across languages and transfer to a different dataset.

\section*{Limitations}
Our analysis is restricted to a single model, Qwen3-4B, and to its non-thinking mode. Recent studies demonstrate that either more parameters \citep{xiong2024can, li-etal-2025-representations} and chain-of-thought reasoning \citep{podolak-verma-2025-read} mitigate overconfidence. How our findings interact with these aspects would be the object of future work. Regarding Experiment 1, we did perform extensive prompting strategies for the abstention and verbal settings, varying the elicitation prompt only in the numeric confidence case. Methodologically, our Experiment~2 focuses on features that align in the same direction for feature activation and logit attribution, underestimating features that diverge across the two (e.g.\ ones that activate in certainty contexts yet push toward uncertainty tokens). Disentangling these two aspects could lead to more fine-grained analysis of the mechanism for certainty and uncertainty expression in controlled stories. Finally, our findings for Experiment 2 inherit the limitations of interpreting models through transcoder features \citep{hanna-etal-2025-circuit, lindsey2025biology}.


\bibliography{custom}

\appendix\label{Appendix}

\section{Experiment 1}\label{ap:dataset_prompts}
Qwen3-4B is deployed using greedy decoding and in the non-thinking mode (with empty \texttt{\textless think\textgreater\textless/think\textgreater}). The model is used through the Huggingface Transformer library\footnote{\href{https://huggingface.co/Qwen/Qwen3-4B}{Qwen3-4B}, \href{https://huggingface.co/mwhanna/Qwen3-4B-transcoders}{Qwen3-4B-transcoders}}.

Regarding prompting, in all settings a marker \texttt{**} has been employed to force the model respecting the expected format. The three settings differ only in the system instruction and in the assistant prefill; within a setting, the certainty and uncertainty scenarios share the same story and question and differ by a single premise.

\subsection{Verbal Prompt}
The prompt is straighforward, with the assistant subject prefill (\texttt{The stamp **}) triggering the generation of certainty (e.g., \emph{is, must}) or uncertainty (e.g., \emph{could, might, may}).

\begin{tcolorbox}[
    colback=white,
    colframe=chineseblack,
    colbacktitle=chineseblack,
    coltitle=white,
    fonttitle=\small\bfseries,
    title=Certainty scenario,
    halign title=center,
    boxrule=1.2pt,
    arc=4mm,
    left=2mm,
    right=2mm,
    top=1mm,
    bottom=1mm
]
\footnotesize
\textbf{System.}~Answer after the \texttt{**}, according to the available information.\\[3pt]
\textbf{User.}~There are three cases in a room: a yellow case, a white case and a black case. A stamp is hidden in one of these cases. It is known that the stamp is not in the white case and not in the black case. Is the stamp in the yellow case?\\[3pt]
\textbf{Assistant (prefill).}~\texttt{\textless think\textgreater\textless/think\textgreater}\quad The stamp \texttt{**}
\end{tcolorbox}
\begin{tcolorbox}[
    colback=white,
    colframe=lightbluecobalt,
    colbacktitle=white,
    coltitle=lightbluecobalt,
    fonttitle=\small\bfseries,
    title=Uncertainty scenario,
    halign title=center,
    boxrule=1.2pt,
    arc=4mm,
    left=2mm,
    right=2mm,
    top=1mm,
    bottom=1mm
]
\footnotesize
\textbf{System.}~Answer after the \texttt{**}, according to the available information.\\[3pt]
\textbf{User.}~There are three cases in a room: a yellow case, a white case and a black case. A stamp is hidden in one of these cases. It is known that the stamp is not in the black case. Is the stamp in the yellow case?\\[3pt]
\textbf{Assistant (prefill).}~\texttt{\textless think\textgreater\textless/think\textgreater}\quad The stamp \texttt{**}
\end{tcolorbox}

\subsection{Abstention Prompt}
The system instruction triggers a typical abstention expression ``I don't know'' for uncertainty expression.
\begin{tcolorbox}[
    colback=white,
    colframe=chineseblack,
    colbacktitle=chineseblack,
    coltitle=white,
    fonttitle=\small\bfseries,
    title=Certainty scenario,
    halign title=center,
    boxrule=1.2pt,
    arc=4mm,
    left=2mm,
    right=2mm,
    top=1mm,
    bottom=1mm
]
\footnotesize
\textbf{System.}~Answer according to the available information. Answer `I don't know' if the answer cannot be determined.\\[3pt]
\textbf{User.}~There are three cases in a room: a yellow case, a white case and a black case. A stamp is hidden in one of these cases. It is known that the stamp is not in the white case and not in the black case. Is the stamp in the yellow case?\\[3pt]
\textbf{Assistant (prefill).}~\texttt{\textless think\textgreater\textless/think\textgreater}\quad \texttt{**}
\end{tcolorbox}

\begin{tcolorbox}[
    colback=white,
    colframe=lightbluecobalt,
    colbacktitle=white,
    coltitle=lightbluecobalt,
    fonttitle=\small\bfseries,
    title=Uncertainty scenario,
    halign title=center,
    boxrule=1.2pt,
    arc=4mm,
    left=2mm,
    right=2mm,
    top=1mm,
    bottom=1mm
]
\footnotesize
\textbf{System.}~Answer according to the available information. Answer `I don't know' if the answer cannot be determined.\\[3pt]
\textbf{User.}~There are three cases in a room: a yellow case, a white case and a black case. A stamp is hidden in one of these cases. It is known that the stamp is not in the black case. Is the stamp in the yellow case?\\[3pt]
\textbf{Assistant (prefill).}~\texttt{\textless think\textgreater\textless/think\textgreater}\quad \texttt{**}
\end{tcolorbox}

\subsection{Numeric Confidence Score Prompt}
\begin{tcolorbox}[
    colback=white,
    colframe=chineseblack,
    colbacktitle=chineseblack,
    coltitle=white,
    fonttitle=\small\bfseries,
    title=Certainty scenario,
    halign title=center,
    boxrule=1.2pt,
    arc=4mm,
    left=2mm,
    right=2mm,
    top=1mm,
    bottom=1mm
]
\footnotesize
\textbf{System.}~Answer a confidence score from 1 (uncertain) to 5 (certain) about how confident you are about the answer to the yes/no question, according to the available information.\\[3pt]
\textbf{User.}~There are three cases in a room: a yellow case, a white case and a black case. A stamp is hidden in one of these cases. It is known that the stamp is not in the white case and not in the black case. Is the stamp in the yellow case?\\[3pt]
\textbf{Assistant (prefill).}~\texttt{\textless think\textgreater\textless/think\textgreater}\quad Confidence: \texttt{**}
\end{tcolorbox}

\begin{tcolorbox}[
    colback=white,
    colframe=lightbluecobalt,
    colbacktitle=white,
    coltitle=lightbluecobalt,
    fonttitle=\small\bfseries,
    title=Uncertainty scenario,
    halign title=center,
    boxrule=1.2pt,
    arc=4mm,
    left=2mm,
    right=2mm,
    top=1mm,
    bottom=1mm
]
\footnotesize
\textbf{System.}~Answer a confidence score from 1 (uncertain) to 5 (certain) about how confident you are about the answer to the yes/no question, according to the available information.\\[3pt]
\textbf{User.}~There are three cases in a room: a yellow case, a white case and a black case. A stamp is hidden in one of these cases. It is known that the stamp is not in the black case. Is the stamp in the yellow case?\\[3pt]
\textbf{Assistant (prefill).}~\texttt{\textless think\textgreater\textless/think\textgreater}\quad Confidence: \texttt{**}
\end{tcolorbox}

\section{Results on the Verbal Setting}

\subsection{Behavioral Overconfidence}
\begin{figure}[h]
  \centering
  \includegraphics[width=\linewidth]{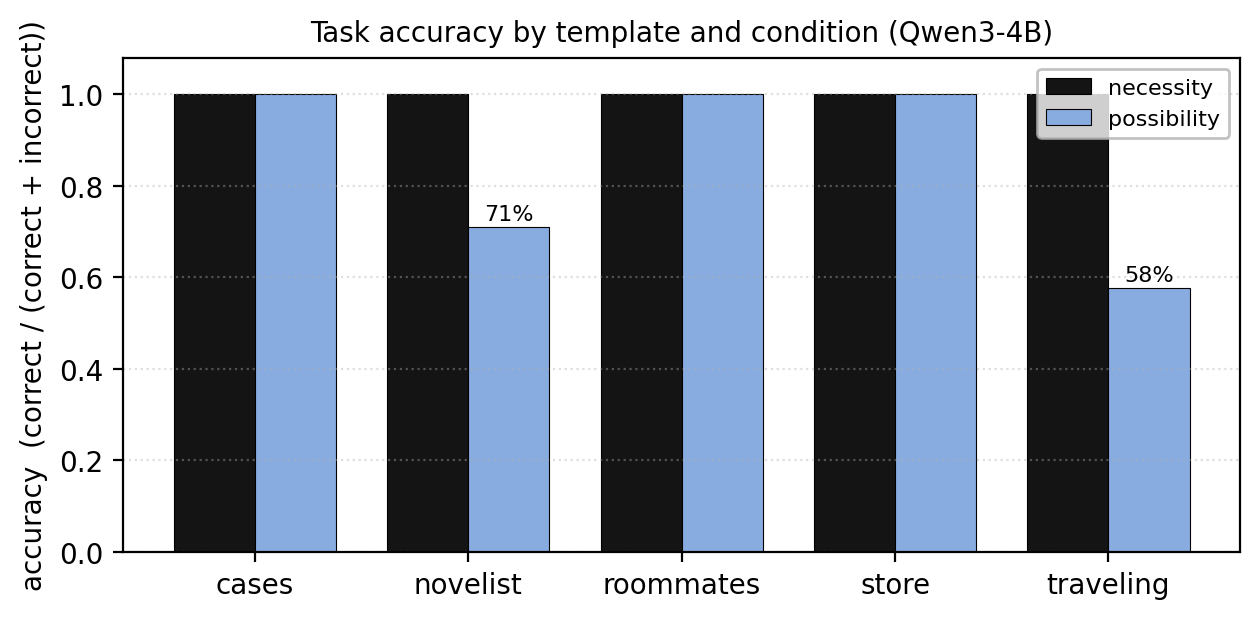}
  \caption{Accuracy by template and condition for the verbal setting.}
  \label{fig:accuracy_by_template}
\end{figure}

We first measure the performance results, with Haiku model annotating the outputs as correct or incorrect. As could be observed in Fig. \ref{fig:accuracy_by_template}, Qwen3-4B answers all 150 necessity prompts correctly (100\%) and 206 of 300 possibility prompts (68.7\%). Of the 94 non-correct possibility responses, 44 are enumerations (the model lists all candidates after the expression of uncertainty), 17 are repetitions of the constraints, and 33 are  errors, i.e., the model emits an assertive \emph{is}/\emph{did} when the correct answer is epistemic. Enumerations and repetitions are not considered in the computation for accuracy. Figure~\ref{fig:accuracy_by_template} shows that accuracy drops are focused on the \textit{novelist} (71\%) and \textit{traveling} (58\%) templates.
\label{sec:results}

In correct scenarios for certainty, it is worth noting the model express certainty through assertive sentences and never with epistemic verbs like \emph{must, should}. In light of these results, the finding by \citet{li-etal-2025-representations} that language models are better in epistemic necessity compared to possibility could be read as an effect of overconfidence.


\subsection{Errors as under-fired possibility circuits}
\begin{figure}[t]
  \centering
  \includegraphics[width=\linewidth]{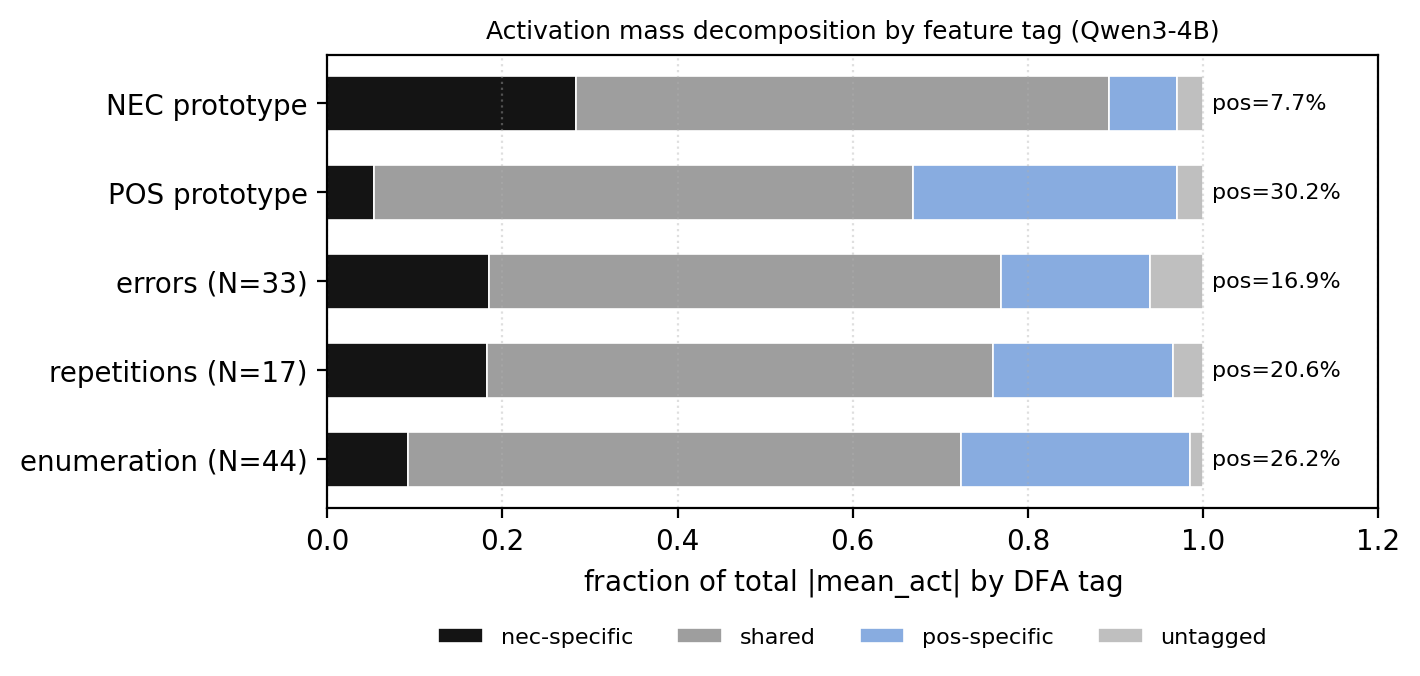}
  \caption{Decomposition of each attribution pool's total $|\mathrm{mean\_act}|$ by DFA tag. Errors under-use the possibility-specific coalition (16.9\% vs.\ 30.2\% in the possibility prototype) and over-use necessity-specific features; enumeration looks like a properly-fired possibility pass.}
  \label{fig:error_profile}
\end{figure}

The three possibility case pools (errors $N{=}33$, repetitions $N{=}17$, enumeration $N{=}44$) are possibility-only. Decomposing each pool's total $|\mathrm{mean\_act}|$ by DFA tag (Figure~\ref{fig:error_profile}) shows that errors stand out from the other two pools: the mass on possibility-specific features drops to 16.9\% (vs.\ 30.2\% in the possibility prototype), while the mass on necessity-specific features rises to 18.5\% (vs.\ 5.3\% in the prototype). Only 1{,}002 of the 2{,}302 possibility-specific features (43.5\%) are even active in the errors attribution pool, and for those that are, their mean activation is 67.8\% of the prototype's ($0.180$ vs.\ $0.265$); on enumeration, by contrast, 63.2\% of possibility-specific features are active, at 105.5\% of prototype activation. Consistently, the mass-weighted similarity verdict flips for errors ($\mathrm{mass}_{\text{on NEC}} = 0.95$ vs.\ $\mathrm{mass}_{\text{on POS}} = 0.92$, closer to the certainty prototype) while enumeration is cleanly closer to possibility ($0.99$ vs.\ $0.89$, ratio $1.12\times$). We read this as under-firing of the sparse possibility override: when the possibility-specific coalition fails to reach its usual activation level, the shared, certainty-tilted default wins and the model emits ``is''.

\subsection{Ablation Intervention}\label{ap:ablation}
\begin{figure}[t]
  \centering
  \includegraphics[width=\linewidth]{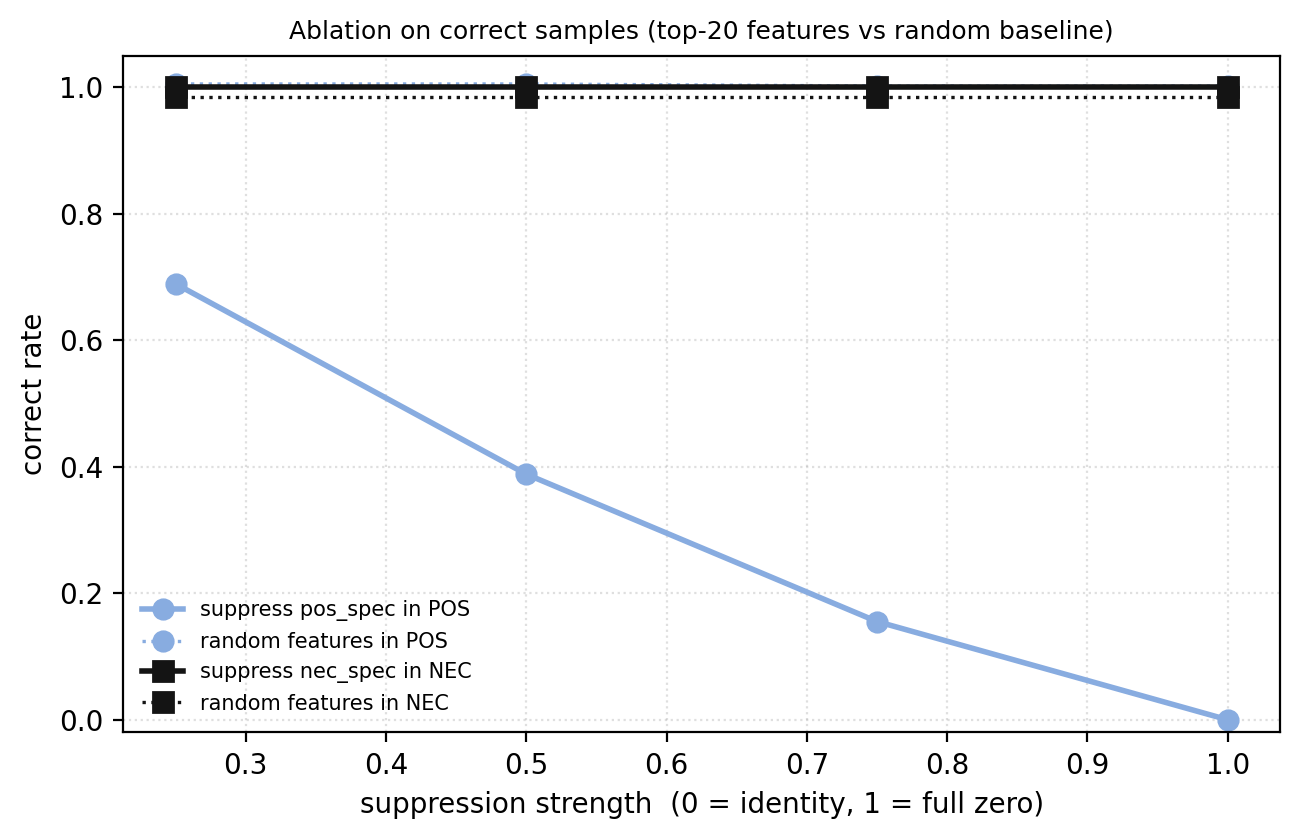}
  \caption{Ablation on correct samples. Progressively suppressing the top-20 possibility-specific features zeroes out possibility accuracy, whereas the symmetric suppression of the top-20 necessity-specific features leaves necessity accuracy at 100\%. Random-feature baselines (dotted) are flat in both conditions.}
  \label{fig:ablation}
\end{figure}

We conversely \emph{suppress} each condition's top-20 own-specific features on correct samples, scaling their activations by $k \in \{0.75, 0.5, 0.25, 0.0\}$, with random-feature baselines of the same size for control (Figure~\ref{fig:ablation}). Suppressing the top-20 possibility-specific features on possibility prompts produces a graded, near-complete collapse of uncertainty expression: correctness falls from 99\% (random baseline) to 69\%, 39\%, 16\%, and 0\% at $k = 0.75, 0.5, 0.25, 0.0$, with $P(\text{epistemic})$ dropping from $0.94$ to $0.02$ and $P(\text{assertive})$ rising to $0.71$---the model switches to confident ``is''/``did'' outputs on all 206 correct possibility samples when the coalition is zeroed. The symmetric experiment on necessity---suppressing the top-20 necessity-specific features on necessity prompts---has \emph{no effect at any level}: 100\% correctness at all four multipliers, with $P(\text{assertive})$ unchanged at $0.80$. Random-feature ablations are flat for both conditions. This dissociation closes the diffuse-vs-sparse story: a small, targeted ablation abolishes the possibility-specific override, but no similarly small ablation can disturb the shared, diffuse certainty default.

\subsection{Feature Visualisation}

\begin{figure*}[h!]
  \centering
  \includegraphics[width=\linewidth]{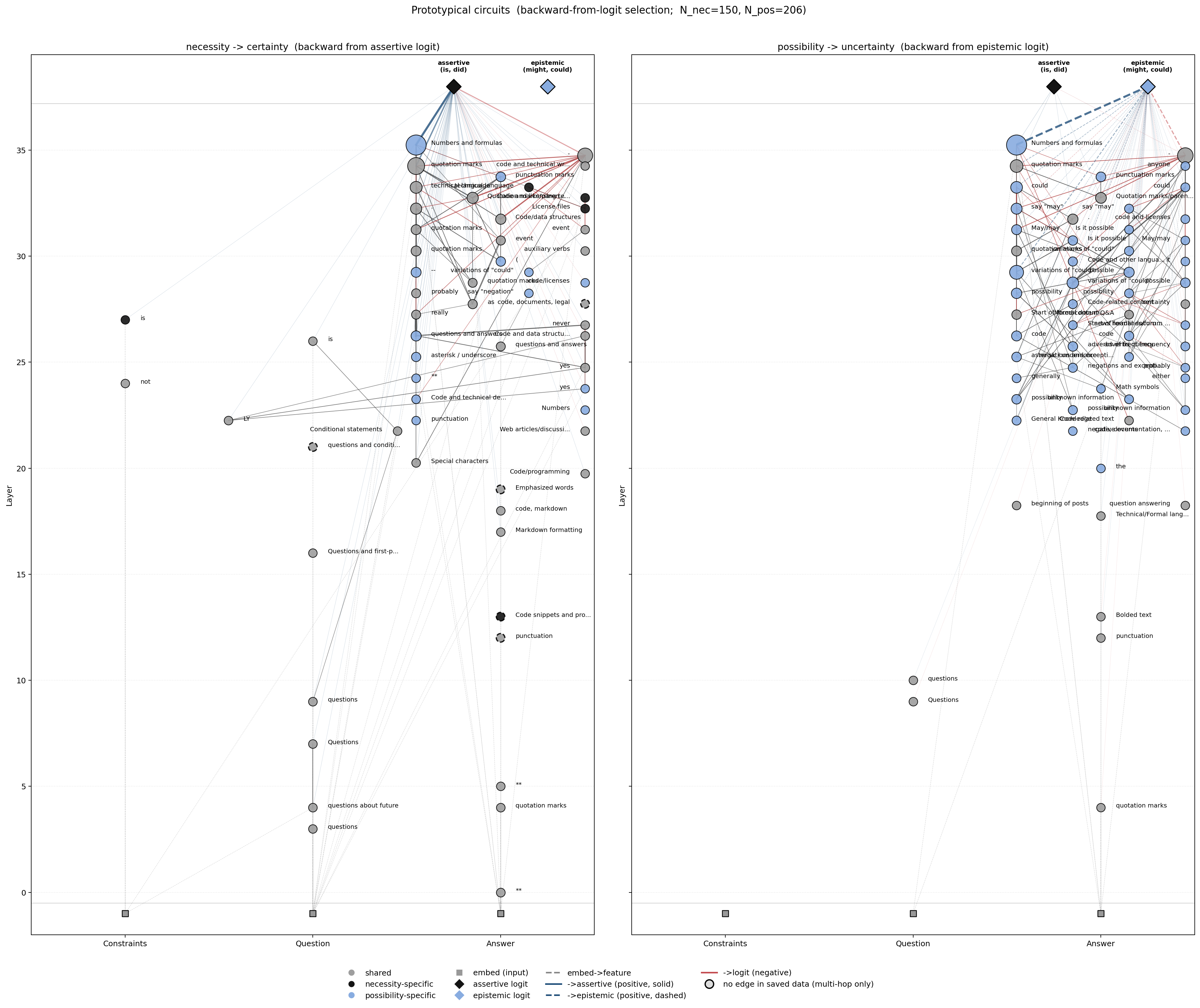}
  \caption{Prototypical circuits for certainty (left) and uncertainty (right) in Qwen3-4B, under a backward-from-logit node selection. Columns are the Constraints / Question / Answer zones; rows are model layers. Nodes are colored by statistical tag (shared, necessity-specific, possibility-specific) and sized by $|\mathrm{mean\_inf}|$; solid and dashed edges go to the assertive and epistemic logit targets respectively, colored by sign. Certainty is carried by a broad distribution of features across layers 22--29, whereas uncertainty concentrates on a sparse set of strongly activated possibility-specific features in the same band.}
  \label{fig:prototype_logits}
\end{figure*}

Reading the logit-ranked panels of Figure~\ref{fig:prototype_logits} by input zone reveals where the middle-layer writers are fed from. In the \emph{certainty} prototype, the \emph{Constraints} zone contributes a small but consistent coalition of lexical-syntactic features labelled ``not'' and ``is'' \footnote{Feature names come from \texttt{Neuronpedia} \citep{neuronpedia}.}. Among these features, present at least in 50\% of the certainty samples, \texttt{L27/F67512} (``is'', $d = +0.33$, $p_\mathrm{adj} = 6\mathrm{e}{-15}$) is robustly certainty-specific under feature attribution. The \texttt{L27} ``is'' feature alone has the largest direct weight to the assertive logit of all Constraints-zone features ($+0.031$ in certainty vs.\ $+0.021$ in uncertainty) and activates with selection frequency $0.98$ in certainty but only $0.40$ in uncertainty---a plausible ``fully-determined'' signal, lit when the negated constraints leave exactly one candidate and the copular ``is'' is the right continuation. In the \emph{uncertainty} prototype, zero Constraints-zone features survive the top-60 logit-rank cut; instead, the influential features concentrate in the \emph{Answer} zone at layers 22--35 and are overwhelmingly uncertainty-specific, with Neuronpedia annotations clustering around epistemic lexicalisation (``variations of \emph{could}'', ``\emph{May/may}'', ``\emph{possibility}'', ``\emph{possible}'', ``say \emph{may}'', ``\emph{Is it possible}''). The coalition is sparse: the top-15 possibility-specific features alone cover 50\% of the signed direct push to the epistemic logit. The last two to three layers are dominated by shared features (quotation marks, technical-language and syntactic markers) that push positively on both logits, so the two circuits converge on the same final-layer writers and the condition-specific decision is made upstream; a minority of shared features flip sign across the two logits and appear as \emph{suppressors} in the logit-ranked panels (e.g.\ \texttt{L35/F123612}, which contributes $-2.83$ to assertive and $-3.44$ to epistemic, damping whichever writer is currently dominant).

\section{Multilingual Experiments}\label{ap:multilingual}

We further extend the controlled stories setting to a multilingual setting, in Chinese and Italian languages. This experiment broadens the research questions, answering the follow-up questions: 
\begin{itemize}
    \item Is the Qwen3-4B overconfidence behavior English-specific or persist across language? 
    \item Do the uncertainty features found for verbal uncertainty generalise across languages?
\end{itemize}

Controlled stories (verbal setting) both for Chinese and Italian were translated by Claude Haiku. As for the English language (Experiment 1), certainty and uncertainty labels for the Qwen3-4B responses are provided using Haiku as a judge.

\begin{figure}[h!]
  \centering
  \includegraphics[width=\linewidth]{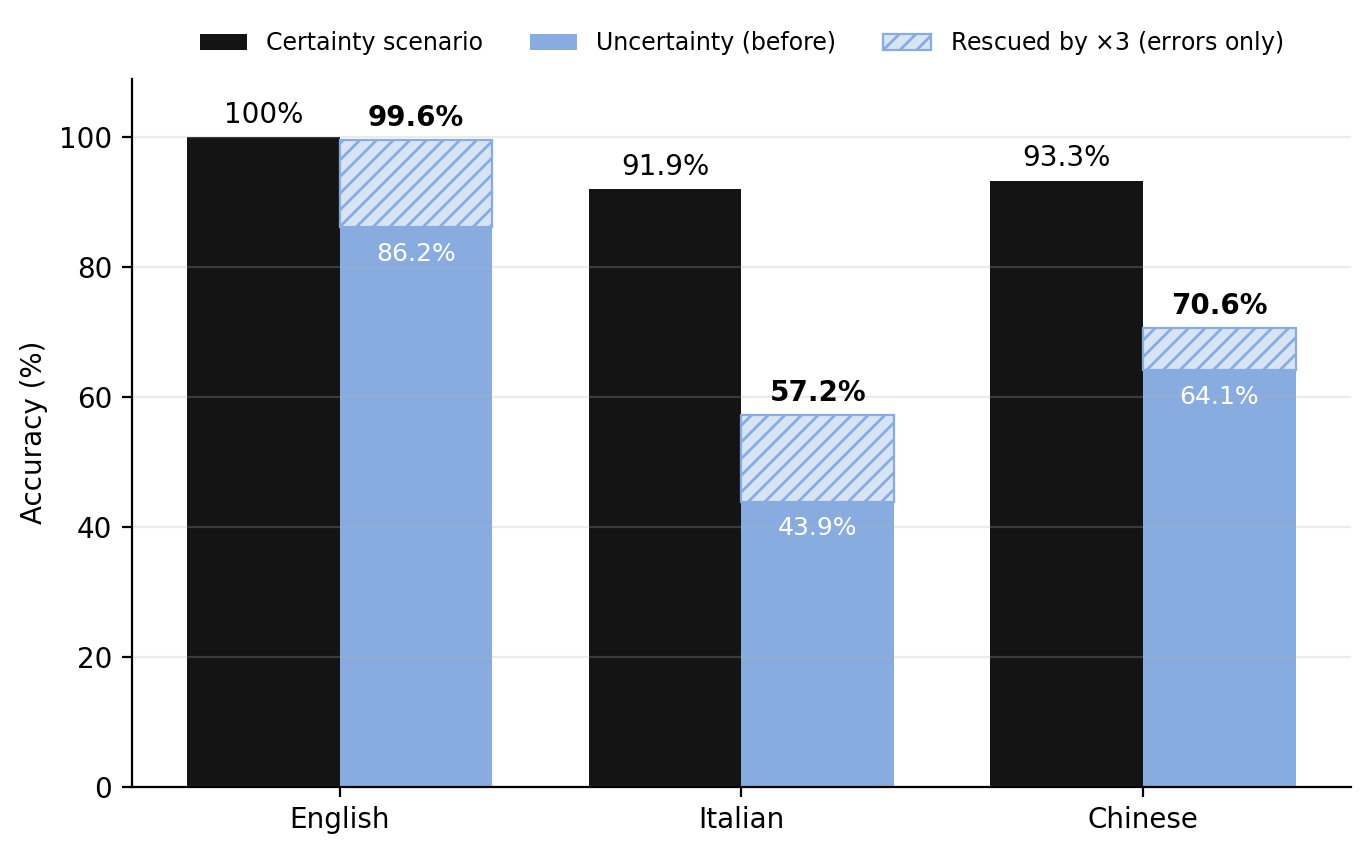}
  \caption{Cross-lingual transfer of the $\times 3$ uncertainty-feature boost, on errors only (following Section \ref{sec:intervention}).}
  \label{fig:multilingual_rescue}
\end{figure}

As reported in Figure~\ref{fig:multilingual_rescue}, Qwen3-4B achieves higher accuracy in certainty scenarios in Chinese and Italian (with accuracy > $90\%$ for both languages), and still exhibiting the overconfidence behaviour observed in uncertainty scenarios ($44\%$ and $64\%$ accuracy respectively). The error rates possibly mirror the multilinguality of Qwen3-4B pretraining data, with English getting best accuracy, followed by Chinese, and then Italian. Applying intervention on the top-20 uncertainty features found for English in the Chinese and Italian datasets results in improved accuracy from $43.9\% \to 57.2\%$ in Italian and $64.1\% \to 70.6\%$ in Chinese for errors (dashed bars in Figure~\ref{fig:multilingual_rescue}). This effect leads to the conclusion that the identified features are abstract uncertainty signal upstream of language-specific lexicalisation, rather than a token-lookup of English modals. These findings extend the linear dimension of uncertainty expression recently found by \citet{ji-etal-2025-calibrating} at the multilingual level.


\section{Implementation Details \& AI Assistance}

The experiments with Qwen3-4B have been run on two Nvidia GeForce RTX 3090
24GB GPU, taking approximately 72 hours in total (12 hours x 6 settings).

AI tools have been employed in this study: Claude Code to draft some parts of the experiments; Chatgpt to provide grammatical corrections and rephrasing of short passages. 

\end{document}